\documentclass[conference]{IEEEtran}
\IEEEoverridecommandlockouts
\usepackage{cite}
\usepackage{amsmath,amssymb,amsfonts}
\usepackage{algorithmic}
\usepackage{graphicx}
\usepackage{textcomp}
\usepackage{xcolor}
\usepackage{booktabs}
\usepackage{hyperref}
\def\BibTeX{{\rm B\kern-.05em{\sc i\kern-.025em b}\kern-.08em
    T\kern-.1667em\lower.7ex\hbox{E}\kern-.125emX}}

\begin{document}

\title{Multi-Model Synthetic Training for Mission-Critical Small Language Models 
}

\author{\IEEEauthorblockN{Nolan Platt$^*$}
\IEEEauthorblockA{\textit{Virginia Tech} \\
Blacksburg, Virginia \\
nolanplatt@vt.edu}
\and
\IEEEauthorblockN{Pragyansmita Nayak}
\IEEEauthorblockA{\textit{Hitachi Vantara Federal} \\
Reston, Virginia\\
pragyan.nayak@hitachivantarafederal.com}
}

\maketitle
\begingroup\renewcommand\thefootnote{*}
\footnotetext{Work done during an internship at Hitachi Vantara Federal.}
\endgroup
\begin{abstract}
Large Language Models (LLMs) have demonstrated remarkable capabilities across many domains, yet their application to specialized fields remains constrained by the scarcity and complexity of domain-specific training data. We present a novel approach that achieves a 261x cost reduction for maritime intelligence by using LLMs as one-time teachers rather than using them directly for inference. Our method transforms 3.2 billion Automatic Identification System (AIS) vessel tracking records into 21,543 synthetic question and answer pairs through multi-model generation (GPT-4o and o3-mini), preventing overfitting and ensuring accurate reasoning. The resulting fine-tuned Qwen2.5-7B model achieves 75\% accuracy on maritime tasks, while being substantially cheaper than using a larger model for inference. We show that smaller, cheaper models - when fine tuned properly - can provide similar accuracy compared to larger models that are prohibitively expensive. Our work contributes to the growing field of synthetic dataset generation for specialized AI applications and presents a highly reproducible framework for domains where manual annotation is infeasible. Beyond expanding research in the growing field of specialized small language models, our approach has immediate applications in maritime safety, security operations, and vessel traffic management systems in various industries.

\end{abstract}


\section{Introduction}
In recent years, Large Language Models (LLMs) have proven successful across diverse natural language tasks, but their usage for specialized domains faces a large challenge: the cost of continuous LLM inference, often reaching thousands of dollars per day for real-time systems \cite{b1}. While Small Language Models (SLMs) could dramatically reduce these costs, they require domain-specific training data that is expensive and time-consuming to manually create and annotate. This limitation is particularly seen in areas like maritime intelligence, where the gap between raw transceiver data and useful insights requires deep expertise that current models simply do not have. Even more so, these mission-critical areas place precision and accuracy at the foremost importance; thus, it is challenging to construct and utilize datasets that can both be trusted and reproducible.

The maritime domain presents a compelling case for this research. The Automatic Identification System (AIS) generates one of the world's largest historical datasets, with vessels broadcasting positional and directional data multiple times per minute. In just 2024 alone, the United States Coast Guard and National Oceanic and Atmospheric Administration (NOAA) collected over 3.2 billion raw AIS data points \cite{b2}, representing comprehensive coverage of vessel movements in U.S. waters. Despite this large amount of data, no comprehensive training datasets exist specifically for language models, preventing them from being trained on how to effectively reason about maritime patterns, detect anomalous behavior, or provide situational awareness - capabilities that are all essential for maritime safety, security, and awareness. 

The creation of such datasets through traditional annotation methods faces considerable roadblocks. Expert analysis of maritime data requires not just domain knowledge, but also computational analysis of vessel trajectories, speed calculations, and pattern recognition across temporal and spatial contexts. A single question about vessel behavior could require analyzing hundreds of data points across multiple vessels and time periods. The cost and complexity of manual annotations at this scale renders traditional, manual approaches infeasible. 

However, recent advances in synthetic data generation offer a potential solution. While LLMs have demonstrated the ability to generate high-quality training data for various tasks \cite{b3}, their application to specialized operational domains remains largely unexplored. We recognize that while deploying large models for real-time maritime analysis can be extremely expensive, likely costing thousands of dollars per day for continuous monitoring, we can leverage these models \textit{once} to generate comprehensive training data for fine-tuning smaller, more efficient models.

This approach fundamentally challenges the modern economics of specialized AI deployment. Rather than choosing between expensive LLM inference or costly manual annotation, we use LLMs as a one-time investment to accelerate domain-specific capabilities in smaller, much cheaper models. Our fine-tuned Qwen2.5-7B model, running on a single NVIDIA H100 GPU, provides maritime intelligence at 0.38\% of GPT-4o's inference cost - a 261x reduction - while achieving 75\% accuracy on domain-specific tasks.

In this work, we present a reproducible framework for generating large-scale training data sets that allow the deployment of cost-effective maritime intelligence systems. Our approach leverages state-of-the-art language and reasoning models: GPT-4o and o3-mini, respectively \cite{b27, b28}. We use these models to transform raw AIS data into linguistically diverse question-answer pairs, followed by supervised fine-tuning (SFT) of smaller models: Magistral Small (24B), Llama 3.1 (8B), and finally Qwen2.5 (7B). We address three critical challenges: \textbf{(1)} efficiently sampling relevant vessel data from billions of records, \textbf{(2)} generating diverse, accurate questions that cover maritime intelligence tasks, and \textbf{(3)} ensuring the synthetic data produces models that match or exceed the performance of larger counterparts on domain-specific tasks.

A critical concern in synthetic data generation is preventing overfitting to the training dataset's specific patterns and biases \cite{b12}. Prior work has shown that models trained exclusively on synthetic data from a \textit{single} LLM can inherit and amplify that model's limitations \cite{b7, b12}. To address this, we employ a multi-model generation strategy, alternating between GPT-4o and o3-mini every seven contexts. This approach introduces reasoning diversity into our dataset, as these models exhibit different generation patterns and problem-solving approaches, helping our fine-tuned model generalize beyond any single LLM's characteristics.

\section*{Key Contributions}
In summary, the key contributions of this paper are as follows:

\textbf{1. Synthetic Maritime Intelligence Dataset}
We transform 3.2 billion AIS records into 21,543 high-quality Q\&A pairs averaging 73,821 tokens each: the first public dataset of its kind for maritime AI. 

\textbf{2. 261x Cost Reduction}
We present a highly reproducible framework for tuning SLMs to highly specialized domains, especially those without an abundance of training data. Our framework reduces annual inference costs from \$2.19M (GPT-4o) to \$8,400 (self-hosted 7B model) while maintaining 75\% accuracy.

\textbf{3. Multi-Model Generation Strategy}
We demonstrate that alternating between generation models prevents overfitting and improves generalization in synthetic datasets.

\textbf{4. Future Outlook for Specialized SLMs}
We provide valuable research that shows that the future of language models may be centered around a series of smaller, fine-tuned SLMs, rather than a singular, expensive LLM. We provide further insight into potential ways to further our work, including neurosymbolic AI and agentic models.

The remainder of this paper is structured as follows: Section \ref{background} reviews related work in synthetic dataset generation and maritime AI applications. Section \ref{methodology} details our methodology for sampling AIS data and prompt engineering. Section \ref{fineTuning} describes our experimental setup and fine-tuning process. Section \ref{results} presents our results on both dataset quality and model performance. Section \ref{discussion} discusses implications for cost-effective AI deployment and limitations. Section \ref{conclusion} concludes with future directions for synthetic data generation in specialized domains.

\section{Background} \label{background}

The development of our AIS-focused SLM builds on three areas of research: synthetic dataset creation for language models, domain-specific fine-tuning, and general maritime AI applications. 

\subsection{Synthetic Dataset Creation}
Synthetic training data has emerged as a solution for specialized domains lacking sufficient annotated datasets. DataDreamer \cite{b3} provides the first comprehensive framework for synthetic data generation, addressing the lack of annotated data for highly specific domains. Liu et al. \cite{b6} present best practices for synthetic data generation, covering applications in reasoning, tool use, and multi-model generation. They provide specific techniques with regard to prompt engineering and retrieval-augmented pipelines.

Despite this, challenges remain in synthetic data quality and model overfitting. Li et al. \cite{b7} identify that subjectivity negatively correlates with synthetic data effectiveness, conducting systematic evaluation across classification tasks. This finding is particularly relevant to our maritime domain, where objective vessel tracking data may be more useful for synthetic generation than subjective maritime assessments.

\subsection{Domain-Specific Fine Tuning}
Domain-specific fine-tuning has proven highly effective across various fields. BioBERT \cite{b4} demonstrated significant improvements on biomedical tasks through domain-specific pre-training, achieving +12.24\% accuracy gains. Similarly, BloombergGPT \cite{b5} showed that a 50B parameter model trained with a mixed approach (363B financial + 345B general tokens) maintains general capabilities while gaining substantial domain expertise.

Recent work by Cheng et al. \cite{b8} provides insights particularly relevant to our approach. Their research reveals that raw domain pre-training can actually \textit{hurt} prompting ability, instead proposing to transform corpora into reading comprehension format. This finding directly influenced our methodology for converting AIS data into question-answer pairs, ensuring the model maintains reasoning capabilities while gaining maritime expertise.

\subsection{Maritime AI Applications}
While machine learning models have achieved success in AIS trajectory prediction \cite{b9} and language models have been applied to maritime domain questions \cite{b10,b11}, the challenge of using language models to directly analyze and reason about raw AIS data remains unexplored. While the aforementioned models are related to the maritime domain, they are not language models capable of reasoning on massive amounts of raw AIS data. Our work bridges this gap by creating the first comprehensive dataset for maritime intelligence language modeling.

\section{Methodology} \label{methodology}
Our methodology addresses three core challenges: efficiently sampling from 3.2 billion AIS records, generating diverse Q\&A pairs that cover maritime intelligence tasks, and preventing overfitting through multi-model generation. All code and analysis notebooks are publicly available \cite{b29}.

\subsection{AIS Data Sampling and Processing}
We developed a  sampling approach to extract representative vessel contexts from the 2024 AIS dataset provided by the USCG and NOAA.  Using Pentaho Data Integration \cite{b16}, we built a pipeline to extract, clean, and load 3.2 billion AIS records into a PostgreSQL database. This allowed us to map vessel types, normalize status categories, and prepare the data for supervised fine-tuning.

Once loaded, we split the AIS data into contexts, with each context containing 200-500 vessels with complete positional data, sampled across:
\begin{enumerate}
    \item Geographic regions (East Coast, West Coast, Gulf of Mexico, Great Lakes)
    \item Port areas versus open water
    \item Diverse time periods (peak/off-peak hours, different seasons)
    \item Vessel types and traffic densities
\end{enumerate}

This stratification ensures our dataset captures the full diversity of maritime operations, from congested port approaches to open-ocean transits.

\subsection{Synthetic Q\&A Generation Pipeline}
Using DataDreamer \cite{b3}, we generated 21,543 Q\&A pairs through a carefully designed multi-model approach. Table \ref{tab:dataset_summary} provides a summary of our dataset.

\begin{table}[htbp]
\centering
\caption{Dataset Summary}
\label{tab:dataset_summary}
\begin{tabular}{ll}
\toprule
Metric & Value \\
\midrule
Total Q\&A Pairs & 21,543 \\
Training Set & 19,389 (90\%) \\
Validation Set & 2,154 (10\%) \\
Average Context Length & 73,821 tokens \\
Source AIS Records & 3.2 billion \\
Generation Models & GPT-4o, o3-mini \\
\bottomrule
\end{tabular}
\end{table}

The complete dataset is publicly available \cite{b30}. Each context generated 12 questions across six diverse categories:
\begin{enumerate}
    \item \textbf{Trajectory Prediction} (3 questions): Future vessel positions
    \item \textbf{Movement Analysis} (2 questions): Speed and heading changes
    \item \textbf{Vessel Counting} (2 questions): Traffic density metrics
    \item \textbf{Data Analysis} (2 questions): Statistical summaries
    \item \textbf{Pattern Detection} (2 questions): Behavioral patterns
    \item \textbf{Anomaly Detection} (1 question): Unusual activities
\end{enumerate}
Example queries include: \textit{"Which vessels near Port of Los Angeles changed course by more than 45 degrees in the past hour?"} (Movement Analysis), and \textit{"Identify any cargo ships exceeding 25 knots within 5 nautical miles of San Francisco Bay"} (Anomaly Detection).

To ensure linguistic diversity and prevent overfitting, we randomized five linguistic styles: Technical/Analytical, Operational/Command, Investigative, Practical User, and Conversational. This variation helps the model generalize across different user types and query variations.

\subsection{Multi-Model Generation Strategy}

Following recent work on synthetic data risks \cite{b12}, we employed a multi-model generation strategy to prevent overfitting to any single model's biases:
\begin{itemize}
    \item \textbf{GPT-4o}: 1,500 contexts (85.7\%)
    \item \textbf{o3-mini}: 250 contexts (14.3\%)
\end{itemize}

We alternated between models every seven contexts, introducing reasoning diversity as these models exhibit different generation patterns and problem-solving approaches. This strategy proved effective, with our fine-tuned model maintaining consistent performance across both source models (75.9\% for GPT-4o-generated vs. 71.4\% for o3-mini-generated questions).

\subsection{Evaluation Methodology}
To comprehensively evaluate our model's performance, we developed a two-part evaluation framework:

\subsubsection{Automatic Evaluation} 

Our primary evaluation method compares numerical values extracted from model responses against correct answers. A response is marked correct if all numerical values (vessel counts, speeds, positions, distances) fall within 10\% of the reference values. This threshold accounts for legitimate variations in calculation methods while ensuring practical accuracy for maritime operations.

\subsubsection{Manual Evaluation}
To validate our automatic pipeline and assess qualitative aspects, we manually evaluated 100 randomly sampled responses. Manual evaluation followed identical numerical accuracy criteria but additionally assessed the correctness of reasoning steps, presence of calculation errors despite correct final answers, and the quality of explanations and maritime domain understanding.

This approach allowed us to validate both the accuracy of our automatic evaluation, as well as the model's reasoning capabilities beyond numerical correctness.

\section{Fine-Tuning and System Architecture} \label{fineTuning}
\subsection{Model Selection Process}
Our initial attempts with Magistral Small (24B) and Llama 3.1 (8B) revealed critical missteps. Magistral memorized patterns without comprehension, while Llama 3.1 hallucinated vessel positions catastrophically. These failures led us to Qwen2.5-7B, selected for its JSON pre-training and native long-context support through YaRN rope scaling \cite{b17}.

\subsection{Training Configuration}
Learning from prior failures, we employed an aggressive training strategy to prevent overfitting while ensuring comprehension.

\subsubsection{Context Extension with YaRN}
The extended context in our model relies on YaRN scaling \cite{b17}, which uses the "NTK-by-parts" interpolation method to extend context windows while maintaining the original quality of the model. YaRN modifies Rotary Position Embeddings (RoPE) \cite{b15} through a frequency-selective approach, as defined by Equations \eqref{gEq} and \eqref{yarnScaling}.

\begin{equation} \label{gEq}
    g(m) = m
\end{equation}

\begin{equation} \label{yarnScaling}
    h(\theta_d) = \left(1 - \gamma(r(d))\right) \frac{\theta_d}{s} + \gamma(r(d))\theta_d.
\end{equation}
The NTK-by-parts interpolation method uses several components that work together to significantly extend contexts:
\begin{itemize}
    \item \textbf{Position preservation:} $g(m) = m$ ensures position indices remain unchanged, unlike position interpolation methods that scale indices directly
    \item \textbf{Base frequency:} $\theta_d = b^{-2d/|D|}$ represents the original RoPE frequency at dimension $d$, with base $b=10000$
    \item \textbf{Scale factor:} $s = 4$ extends our context window from 32k to 131k tokens (a 4× increase)
    \item \textbf{Ramp function:} $\gamma(r(d))$ smoothly transitions from 0 to 1 based on the frequency's wavelength
    \item \textbf{Wavelength ratio:} $r(d) = L/\lambda_d$, where $L$ is the original context length and $\lambda_d = 2\pi b^{2d/|D|}$ is the wavelength at dimension $d$
\end{itemize}

$h(\theta_d)$ selectively modifies frequencies. When $\gamma = 0$, we get full interpolation: $\theta_d/s$. Evidently, when $\gamma = 1$ - at higher frequencies - there is no change: $\theta_d$. This preserves high-frequency information, which is critical for distinguishing nearby tokens.

For AIS data, this is essential. Each context we use contains hundreds of vessel records with similar coordinates, such as:
\begin{verbatim}
{"lat": 23.7022, "lon": -120.9633, ...}
{"lat": 23.7023, "lon": -120.9634, ...}
\end{verbatim}
If we used standard RoPE scaling, all frequencies would be compressed equally by a constant of $s$. This would thus make all adjacent vessels with similar coordinates indistinguishable. As shown, YaRN preserves high-frequency components that differentiate  nearby vessels, while also extending low-frequency components for long-range patterns.

Qwen2.5's pre-training on JSON data provides understanding of structured data, while YaRN maintains precision across a large context window. Llama 3.1 and Magistral lacked both approaches, which likely led to the discussed hallucinations.

\subsubsection{Loss Function and Optimization}
A major issue with synthetic training is the likeliness of overfitting (i.e., memorization - rather than learning - of the training data). As one way to address this, we optimized our training using cross-entropy loss \cite{b22} with label smoothing and one hot encoding, as shown in Equations \eqref{crossEntropy1} and \eqref{crossEntropy2}.

\begin{equation} \label{crossEntropy1}
\mathcal{L} = -\sum_{i=1}^{V} y_i' \log(p_i)
\end{equation}

where the smoothed label distribution is:
\begin{equation} \label{crossEntropy2}
y_i' = (1-\epsilon)y_i + \frac{\epsilon}{V}
\end{equation}

with:
\begin{itemize}
    \item $\epsilon = 0.1$ (smoothing parameter)
    \item $V$ = vocabulary size
    \item $y_i$ = one-hot encoded true label
    \item $p_i$ = predicted probability for token $i$
\end{itemize}

By using label smoothing, we prevent our model from becoming overconfident on our synthetic dataset. More importantly, it helps ensure the model actually generalizes and learns how to \textit{get} to the answers, rather than just memorizing the tokens themselves. This is achieved by assigning small probabilities $(0.1/V)$ to incorrect tokens, rather than simply setting them to zero.

\begin{table}[htbp]
\centering
\caption{Training Hyperparameters for Qwen2.5-7B}
\label{tab:hyperparameters}
\begin{tabular}{ll}
\toprule
\textbf{Parameter} & \textbf{Value} \\
\midrule
Base Model & Qwen2.5-7B-Instruct \\
Context Length & 131k (YaRN factor 4.0) \\
Method & QLoRA \cite{b13} \\
LoRA Rank & 256 \\
LoRA Alpha & 512 \\
Dropout & 0.1 \\
Learning Rate & $2 \times 10^{-4}$ \\
Batch Size & 1 \\
Warmup Steps & 100 \\
Gradient Accumulation & 16 steps \\
Training Hardware & 1x NVIDIA H100 (80GB) \\
Training Duration & 12 hours \\
Final Loss & 0.117 (train) / 0.084 (eval) \\
\bottomrule
\end{tabular}
\end{table}

Critically, we positioned questions \textit{before} vessel data in prompts to prevent truncation at extreme context lengths. While this may seem like a minor change, it proved essential for maintaining accuracy on complex queries requiring full context analysis.

\subsection{System Architecture}
\begin{figure}[htbp]
    \centering
    \includegraphics[width=\columnwidth]{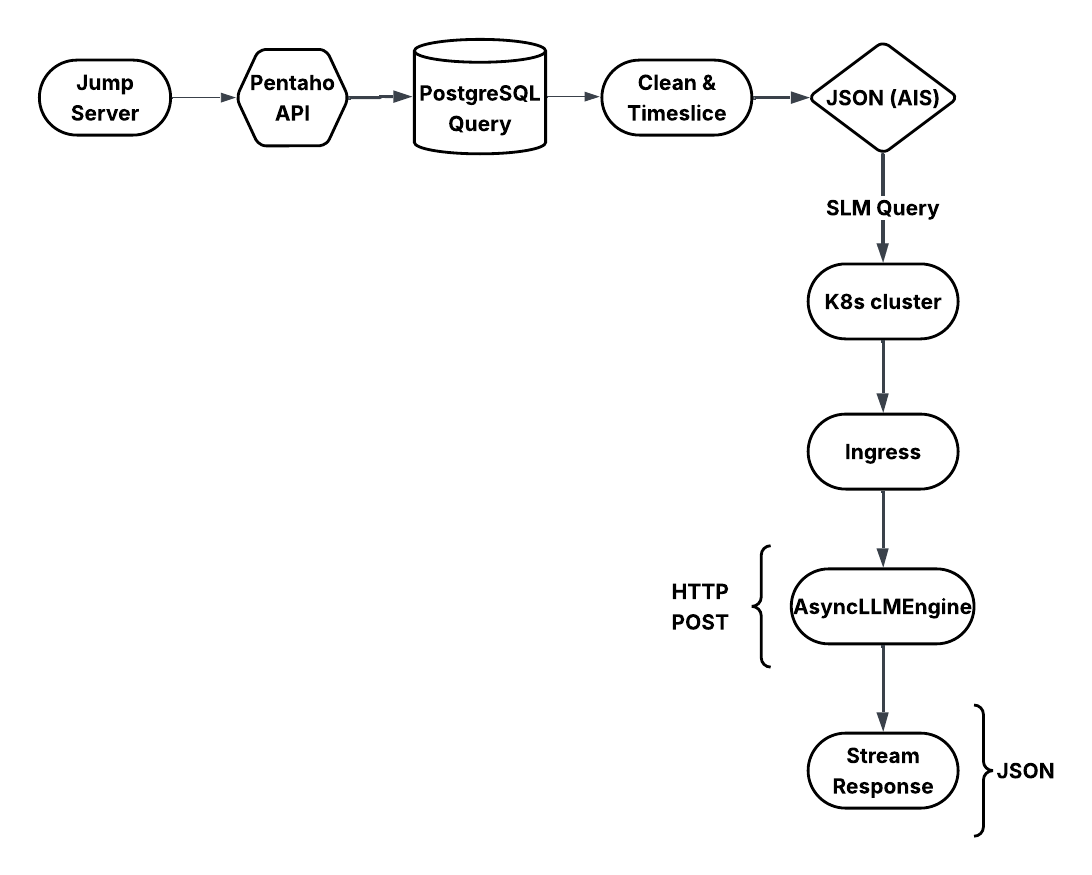}
    \caption{System architecture for real-time maritime intelligence. User queries trigger AIS data retrieval via our Pentaho ETL pipeline, with the fine-tuned model providing streaming JSON responses.}
    \label{fig:system-architecture}
\end{figure}

Figure \ref{fig:system-architecture} illustrates our deployed system architecture, designed for real-time maritime intelligence at scale.

The system processes queries through:
\begin{enumerate}
    \item Query parsing and temporal/spatial extraction
    \item PostgreSQL retrieval of relevant AIS records
    \item Context assembly with vessel data formatting
    \item Inference through our fine-tuned Qwen2.5-7B
    \item Streaming JSON response generation
\end{enumerate}

\section{Results} \label{results}
Our evaluation reveals that traditional NLP metrics fail to show language models' performance in very specialized domains. As shown in Table \ref{tab:metrics_comparison}, despite having near-zero BLEU scores, manual evaluation underlines the model \cite{b31} has strong accuracy and consistency.
\begin{table}[htbp]
\centering
\caption{Traditional Metrics vs. Domain Performance}
\label{tab:metrics_comparison}
\begin{tabular}{llrl}
\toprule
\textbf{Type} & \textbf{Metric} & \textbf{Score} & \textbf{Assessment} \\
\midrule
NLP & BLEU & 0.091\% & Extremely poor \\
NLP & ROUGE-L & 10.9\% & Very poor \\
NLP & BERTScore F1 & -0.183 & Negative correlation \\
\midrule
Domain & Manual Accuracy & 75.0\% & Strong performance \\
Domain & Shows Reasoning & 98.0\% & Near perfect \\
Domain & Avg. Response Length & 2,773 chars & 9.2× verbosity \\
\bottomrule
\end{tabular}
\end{table}

\begin{figure*}[h!]
\centering
\includegraphics[width=0.8\textwidth]{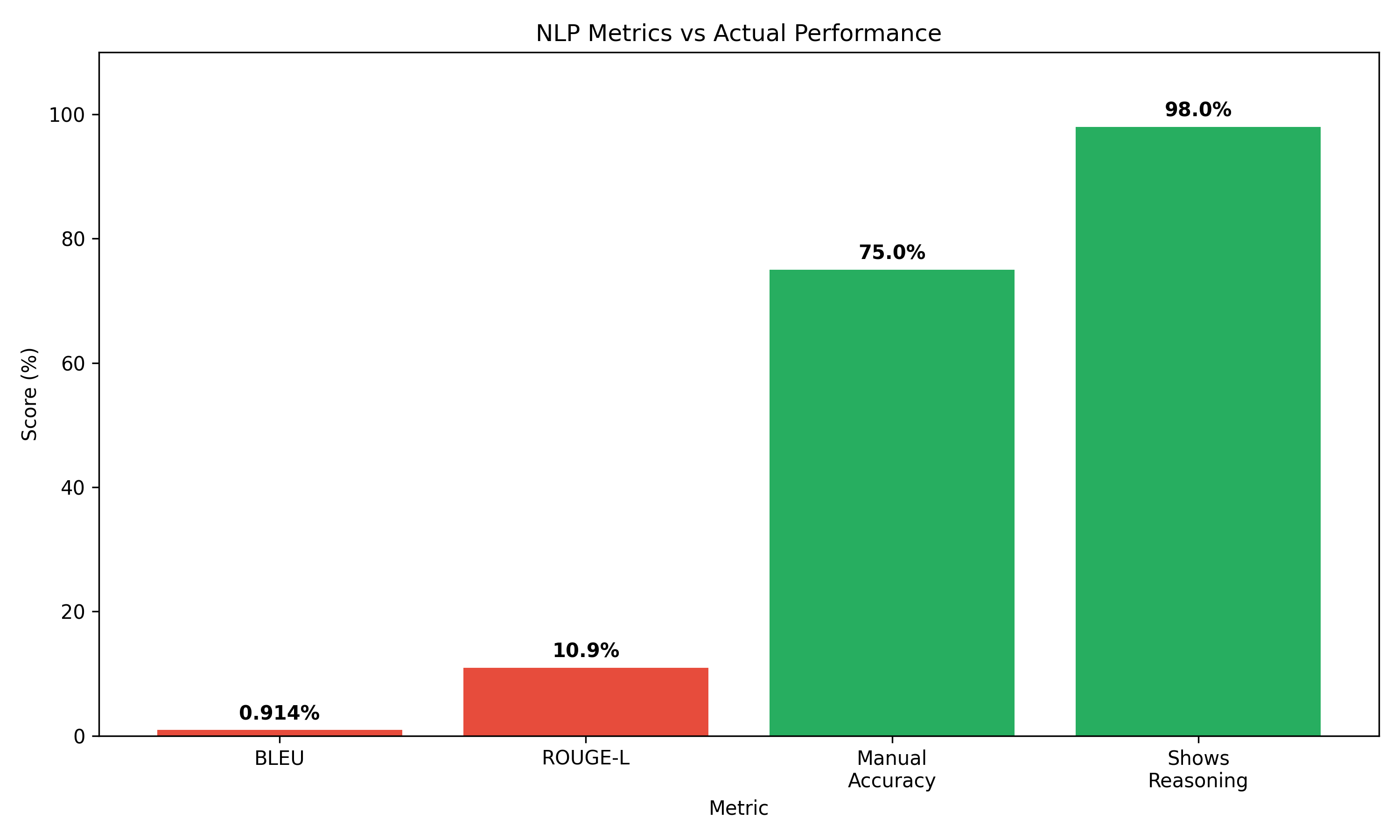}
\caption{Traditional metrics versus actual performance. The disparity between NLP metrics (BLEU, ROUGE-L) and actual performance (accuracy, reasoning) shows the inconsistency of standard NLP evaluation methods for domain-specific tasks.}
\label{fig:metrics}
\end{figure*}

The poor scores for NLP metrics stem from our model's highly verbose, educational responses. Specifically, the model tends to provide extremely detailed explanations rather than trying to match reference answers. Figure \ref{fig:metrics} more clearly shows the disparity between traditional NLP metrics and the actual results. The BLEU score was calculated as shown in Equation \eqref{bleu} with respect to brevity penalty $BP$ in Equation \eqref{brevity}.
\begin{equation} \label{bleu}
\text{BLEU} = BP \cdot \exp\left(\sum_{n=1}^{4} w_n \log p_n\right)
\end{equation}
where $p_n$ is the precision of n-grams:
\begin{equation} \label{brevity}
BP = \begin{cases}
1 & \text{if } c > r \\
e^{1-r/c} & \text{if } c \leq r
\end{cases}
\end{equation}

with $c$ = candidate length and $r$ = reference length. Our 9.2× token increase ($c \gg r$) sets $BP = 1$. However, $p_n$ approaches zero because the detailed explanations of our model do not share exact phrases with the synthetic dataset. Thus, while BLEU provides meaningful insight for NLP, it is likely not the best metric when evaluating models that were trained from multi-model synthetic datasets like ours.

\subsection{Statistical Analysis of Performance}
To validate our evaluations, we compared manual evaluations ($n=100$) with automated evaluations ($n=500$) using a two-proportion z-test, which will tell us if the difference between the two is statistically significant. The generic formula for a two-proportion z-test is shown in Equation \eqref{zTestGeneric}:

\begin{equation}\label{zTestGeneric}
z = \frac{p_1 - p_2}{\sqrt{p_{pool}(1-p_{pool})(1/n_1 + 1/n_2)}}
\end{equation}

We first calculate the pooled proportion using Equation \eqref{pooled}:

\begin{equation} \label{pooled}
p_{pool} = \frac{x_1 + x_2}{n_1 + n_2} = \frac{75 + 354}{100 + 500} = 0.715
\end{equation}

where $x_1 = 75$ correct manual evaluations and $x_2 = 354$ correct automated evaluations.

Next, we calculate the standard error of the difference between proportions, shown in Equation \eqref{SE}:

\begin{equation}\label{SE}
\begin{aligned}
SE &= \sqrt{p_{\text{pool}}\!\left(1-p_{\text{pool}}\right)
          \left(\tfrac{1}{n_{1}}+\tfrac{1}{n_{2}}\right)}
\\[4pt]  
   &= \sqrt{0.715\cdot 0.285\cdot 0.012} = 0.04945 .
\end{aligned}
\end{equation}

With the pooled proportion and standard error calculated, we can now compute the z-statistic using Equation \eqref{zTest}:

\begin{equation}\label{zTest}
z = \frac{p_1 - p_2}{SE} = \frac{0.75 - 0.708}{0.04945} = \frac{0.042}{0.04945} = 0.85
\end{equation}

where $p_1 = 0.75$ represents the manual accuracy and $p_2 = 0.708$ represents the automated accuracy. 

$z=0.85$ corresponds to a p-value of $0.3957$, which indicates no significant difference between the two evaluation methods and thus validates that we are reliably capturing the model's performance.

\subsection{Statistical Confidence Analysis}
As shown in Table \ref{tab:confidence_intervals}, we calculated 95\% confidence intervals using the Wilson score method \cite{b25} to show the reliability of our model between metrics. This method tends to provide more accurate bounds than approximation methods, especially in data with small sample sizes.

\begin{table}[htbp]
\centering
\caption{95\% Confidence Intervals}
\label{tab:confidence_intervals}
\begin{tabular}{lccc}
\toprule
Metric & Point Estimate & 95\% CI \\
\midrule
Manual Accuracy & 0.750 & [0.657, 0.825] \\
Automated Accuracy & 0.708 & [0.667, 0.746] \\
\midrule
\textit{By Question Type} & & \\
Anomaly Detection & 1.000 & [0.646, 1.000] \\
Trajectory Prediction & 0.815 & [0.633, 0.918] \\
Pattern Detection & 0.833 & [0.552, 0.953] \\
Vessel Count & 0.706 & [0.469, 0.867] \\
Data Analysis & 0.652 & [0.449, 0.812] \\
Movement Analysis & 0.615 & [0.355, 0.823] \\
\bottomrule
\end{tabular}
\end{table}
The narrow interval for automated accuracy $[0.667, 0.746]$ validates our evaluation consistency, while the large displacement for anomaly detection $[0.646, 1.000]$ reflects the small sample size ($n=7$) for this category.

\subsection{Performance Across Question Types}
Table \ref{tab:confidence_intervals} shows performance variations across different maritime tasks. Our model achieves perfect accuracy on anomaly detection (100\%, n=7), while struggling with movement analysis (61.5\%, n=13). This difference likely reflects the  difference in complexity of the two: anomaly detection relies on clear threshold violations (e.g., vessel speeds exceeding physical limits), while movement analysis requires interpretation of heading changes and acceleration patterns, requiring both sound physics and pattern identification.

The automated evaluation (n=500) confirms our manual findings, with trajectory prediction maintaining 82.6\% accuracy at scale. The small confidence intervals for large categories (trajectory prediction: $[73.3\%, 89.6\%]$) further validate the reliability of our model.

\subsection{Cost Analysis}
\begin{figure*}[h!]
\centerline{\includegraphics[width=1.1\textwidth]{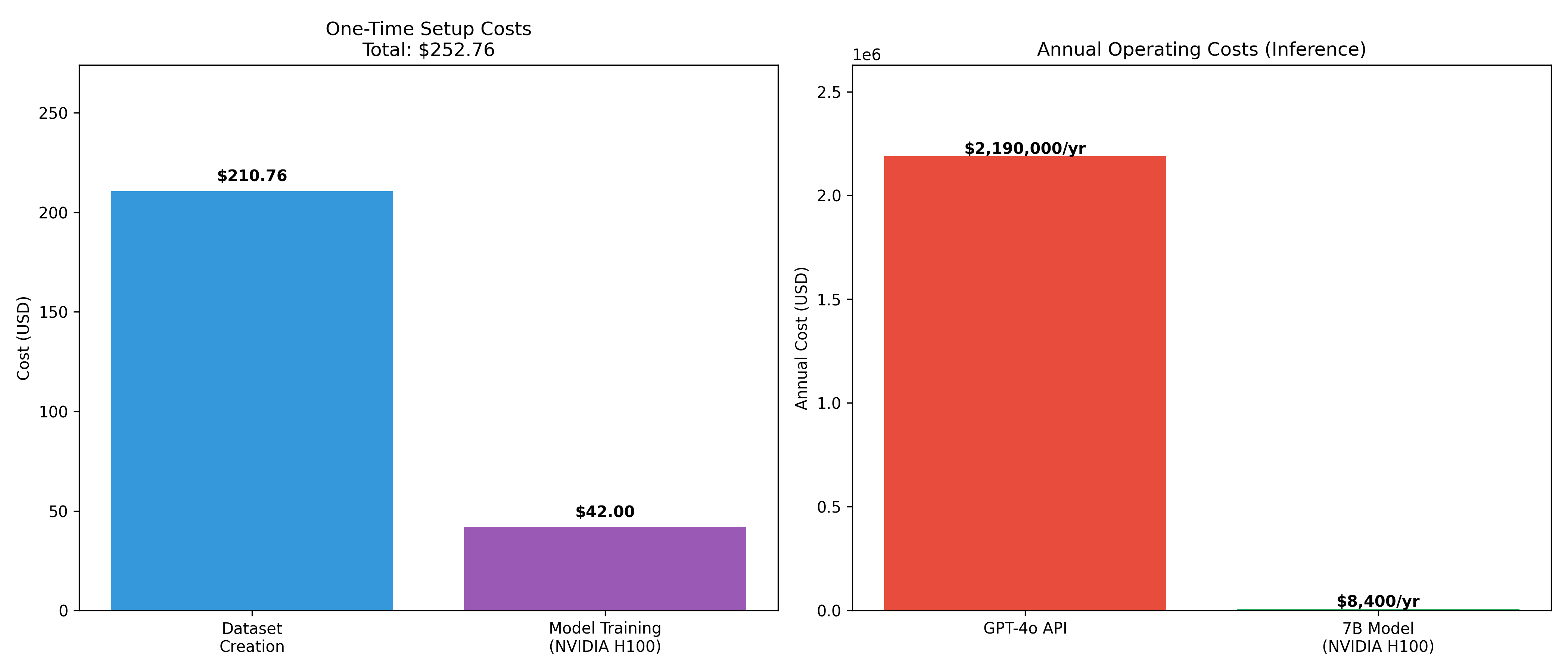}}
\caption{Using a small language model (SLM) is significantly cheaper for domain-specific tasks than relying on larger, more expensive models.\newline Calculations based on $~\approx$ 10,000 queries per day.}
\label{fig:cost}
\end{figure*}

Figure \ref{fig:cost} shows the dramatic economic impact that our approach enables. The 261x cost reduction from \$2.19M to \$8,400 annually fundamentally challenges the cost of deploying specialized AI systems.

\section{Discussion} \label{discussion}
Our results show how specialized AI systems can be developed and deployed cheaper and more efficiently. By addressing each of our key contributions, we show how synthetic data generation, strategic model selection, and unique evaluation approaches can be combined to improve access to domain-specific AI systems.

\subsection{The First Maritime Intelligence Dataset}
Our transformation of 3.2 billion AIS records into 21,543 high-quality Q\&A pairs represents more than just a technical achievement. Rather, it establishes a new standard for domain-specific dataset creation and organization. Each Q\&A pair, averaging 73,821 tokens, has the complexity of real maritime operations, while still maintaining the structured format necessary for effective fine-tuning of language models.

Our dataset's diversity across six different categories allows comprehensive coverage of maritime tasks. Each unique category can contain hundreds of different types of questions. By using a multi-model synthetic approach, and using state of the art language and reasoning models, we attempt to cover all edge cases and specific scenarios in the domain. This distribution reflects real-world priorities: two situations are never the same, and using large, powerful models to generate the dataset helps cover all cases.

Further, our dataset helps bridge the gap between transceiver data and intelligence. Traditional AIS datasets contain positional broadcasts but lack actual questions, answers, or reasoning, thus preventing their usability in  decision-making. By generating questions that require multi-vessel analysis, pattern recognition, and reasoning across temporal and spatial contexts, we create training data that teaches models to actually think like maritime analysts rather than merely process coordinates. 

\subsection{261x Cost Reduction}
We have shown that SLMs make a substantial economic impact on highly-specialized domains. In our case, it extends far beyond just small savings. To query the same questions to GPT-4o, i.e., using our synthetic dataset with the same average number of tokens per context, it would cost about \$2.19M annually. This fundamentally limits its usage to major conglomerates, government agencies, and other organizations with significant spending power. Our approach fundamentally challenges that notion; as shown, our model can provide similar results at just \$8,400 annually, enabling deployment to industries that were previously impossible, including small port authorities, developing nations, and research institutions. Our work provides the framework for efficiently creating small language models for highly specific domains, even in organizations where cost is a limiting factor. 

\subsection{Evaluation Paradox}
Our results expose a paradox in current evaluation in NLP. As shown, models that are optimized for human use, particularly those at scale requiring significant detail, may score poorly on metrics designed for linguistic similarity. While the NLP metrics are very poor for our model, this is not a failure. Instead, it is a failure of the metrics themselves \textit{for this use case}. BLEU's n-gram matching penalizes our model's comprehensive explanations. While BLEU may see zero overlap to reference answers, actual humans likely value the additional context, detail, and reasoning -- especially in mission-critical domains like ours. Our 9.2x verbosity ratio reflects domain expertise, not incorrectness.

\subsection{Multi-Model Generation Strategy}
Our use of both GPT-4o and o3-mini addresses a challenge in synthetic data generation: preventing overfitting and model collapse. The negligible performance difference between the two (75.9\% for GPT-4o-generated vs. 71.4\% for o3-mini-generated) validates this approach.

Beyond these statistical benefits, we informally observed qualitative differences:
\begin{enumerate}
    \item GPT-4o was great at probabilistic trajectory predictions, generating questions about likely vessel paths based on its historical patterns in the data
    \item o3-mini was more focused on rule-based violations, creating questions about regulatory compliance and safety thresholds
\end{enumerate}

This two-fold generation strategy helped our fine-tuned model's reasoning abilities across many different tasks in the domain. For anomaly detection, the model combines both approaches: using statistical methods to identify outliers while also applying numerical methods for safety violations. This approach, which was likely infeasible if we did training from a single source, helped contribute to the model's 100\% accuracy on anomaly detection questions. 

In addition, our strategy is resilient to model end-of-life. As AI providers (e.g. OpenAI in our case) retire models, the synthetic datasets generated from multiple sources maintain their value much longer than those dependent on the reasoning and characteristics of a single model.

\subsection{Future Outlook for Specialized SLMs}
Our work points toward using smaller models as subsets of a larger system. Neurosymbolic approaches such as Scallop \cite{b26} could enhance our maritime model by combining physics constraints with neural pattern recognition, potentially achieving higher precision with significantly less training data.

\subsection{Limitations}
While our results are promising, there are several limitations that should be discussed:

\textbf{Temporal Degradation:} Maritime patterns can evolve with new routes, regulations, and technologies. Our dataset is based on AIS data from 2024; if used in production, the model should be retrained on an annual basis.

\textbf{Geographic Constraints:} As mentioned, we trained exclusively in US waters. This may limit performance in regions outside these regions. We have not tested the model on AIS data outside of the US, so its ability to perform on international data is unpredictable. Deployment in an international region would likely benefit from region-specific fine-tuning. 

\textbf{Vulnerabilities:} Synthetic training cannot capture all AIS manipulation techniques. Criminals with significant expertise in AIS spoofing may be using techniques that our dataset did not capture, and hence our model cannot identify. In high-stakes environments, this likely necessitates a hybrid approach with human analyzers.

\textbf{Context Window:} Despite extending our model's context window to 131k tokens, specific scenarios (major ports during peak hours) could likely exceed limits. This would require strategic sampling techniques based on the situation.

\section{Conclusion} \label{conclusion}
We have shown that specialized language models do not require specific experts, expensive budgets, or high compute power. Our methodology achieves a 261x cost reduction while maintaining 75\% accuracy on maritime intelligence tasks.

The technical innovations we present help provide a reproducible framework for any field where structured data is abundant but expertise and compute power is scarce. Our discovery that traditional NLP metrics fail for specialized applications calls for new evaluation methods focused on operational utility, especially in mission-critical applications.

Most significantly, we have shown that the future of domain-specific language models lies in smaller, cheaper models, rather than one singular large model. Using LLMs as teachers (synthetic data generation), rather than for inference, we help decrease an estimated \$2.19M annual expense to \$8,400, which can allow organizations to use specialized models for domains that were previously infeasible. 

As we look toward the future of neurosymbolic and agentic AI systems, our work provides the foundation for smaller, specialized intelligence. Each specialized domain can develop its own expert models, thus combining to create systems more capable than any single model. We envision an abundance of affordable, specialized models that bring expert analysis to every corner of human intelligence in the world.


\begin{thebibliography}{00}

\bibitem{b1} Y. Xia et al., ``Understanding the Performance and Estimating the Cost of LLM Fine-Tuning,'' arXiv:2408.04693, 2024.
\bibitem{b2} NOAA Office for Coastal Management, ``Nationwide Automatic Identification System 2024,'' U.S. Coast Guard Navigation Center, Feb. 2025.
\bibitem{b3} A. Patel, C. Raffel, and C. Callison-Burch, ``DataDreamer: A Tool for Synthetic Data Generation and Reproducible LLM Workflows,'' in \textit{Proc. ACL 2024}, pp. 3781-3799, 2024.
\bibitem{b4} J. Lee et al., ``BioBERT: a pre-trained biomedical language representation model for biomedical text mining,'' \textit{Bioinformatics}, vol. 36, no. 4, pp. 1234-1240, 2020.
\bibitem{b5} S. Wu et al., ``BloombergGPT: A Large Language Model for Finance,'' arXiv:2303.17564, 2023.
\bibitem{b6} R. Liu et al., ``Best practices and lessons learned on synthetic data for language models,'' arXiv:2404.07503, 2024.
\bibitem{b7} Z. Li et al., ``Synthetic Data Generation with Large Language Models for Text Classification: Potential and Limitations,'' in \textit{Proc. EMNLP 2023}, 2023.
\bibitem{b8} D. Cheng, S. Huang, and F. Wei, ``Adapting Large Language Models via Reading Comprehension,'' in \textit{Proc. ICLR}, 2024.
\bibitem{b9} H. Li, H. Jiao, and Z. Yang, ``AIS data-driven ship trajectory prediction modelling and analysis based on machine learning and deep learning methods,'' \textit{Transportation Research Part E}, vol. 175, p. 103152, 2023.
\bibitem{b10} W. Nguyen et al., ``Llamarine: Open-source Maritime Industry-specific Large Language Model,'' arXiv:2503.00203, 2025.
\bibitem{b11} Zhang et al., ``KUNPENG: An Embodied Large Model for Intelligent Maritime,'' arXiv:2407.09048, 2024.
\bibitem{b12} Gerstgrasser et al., ``Is Model Collapse Inevitable? Breaking the Curse of Recursion by Accumulating Real and Synthetic Data,'' arXiv:2404.11597, 2024.
\bibitem{b13} T. Dettmers et al., ``QLoRA: Efficient Finetuning of Quantized LLMs,'' arXiv:2305.14314, 2023.
\bibitem{b14} J. Su et al., ``RoFormer: Enhanced Transformer with Rotary Position Embedding,'' arXiv:2104.09864, 2021.
\bibitem{b15} S. Wang et al., ``Resonance RoPE: Improving Context Length Generalization of Large Language Models,'' in \textit{ACL 2024 Findings}, arXiv:2403.00071, 2024.
\bibitem{b16} Pentaho Corporation, ``Pentaho Data Integration,'' 2024. [Online]. Available: https://www.hitachivantara.com/en-us/products/pentaho-plus-platform.html
\bibitem{b17} B. Peng et al., ``YaRN: Efficient Context Window Extension of Large Language Models,'' arXiv:2309.00071, 2023.
\bibitem{b18} T. Dao et al., ``FlashAttention: Fast and Memory-Efficient Exact Attention with IO-Awareness,'' arXiv:2205.14135, 2022.
\bibitem{b19} T. Chen et al., ``Training Deep Nets with Sublinear Memory Cost,'' arXiv:1604.06174, 2016.
\bibitem{b20} H. Jin, Y. Wu, et al., ``Rethinking Learning Rate Tuning in the Era of Large Language Models,'' arXiv:2309.08859, 2023.
\bibitem{b21} A. Pareja et al., ``Unveiling the Secret Recipe: A Guide For Supervised Fine-Tuning Small LLMs,'' arXiv:2412.13337, 2024.
\bibitem{b22} J.W. Shim, ``Enhancing cross entropy with a linearly adaptive loss function for optimized classification performance,'' \textit{Scientific Reports}, vol. 14, p. 27405, 2024.
\bibitem{b23} C. An et al., ``Training-Free Long-Context Scaling of Large Language Models,'' arXiv:2402.17463, 2024.
\bibitem{b24} R.S. Raju et al., ``Constructing Domain-Specific Evaluation Sets for LLM-as-a-judge,'' in \textit{ACL CustomNLP4U Workshop}, 2024.
\bibitem{b25} E.B. Wilson, ``Probable Inference, the Law of Succession, and Statistical Inference,'' \textit{Journal of the American Statistical Association}, vol. 22, no. 158, pp. 209-212, 1927.
\bibitem{b26} Z. Li, J. Huang, and M. Naik, ``Scallop: A Language for Neurosymbolic Programming,'' in \textit{Proc. PLDI 2023}, 2023.

\bibitem{b27} Achiam et al. ''GPT-4 Technical Report'' in \textit{arXiv preprint arXiv:2303.08774}, 2023.

\bibitem{b28} OpenAI, ``o3-mini System Card,'' 2025. [Online]. Available: https://cdn.openai.com/o3-mini-system-card-feb10.pdf.

\bibitem{b29} N. Platt and P. Nayak, ``Maritime-SLM-Training: Multi-Model Synthetic Generation and Fine-Tuning Pipeline,'' Figshare, 2025. [Software]. doi: 10.6084/m9.figshare.29709053.v2

\bibitem{b30} N. Platt and P. Nayak, ``AIS QA Dataset: Synthetic Question-Answer Dataset for Maritime Intelligence,'' Figshare, 2025. [Dataset]. doi: 10.6084/m9.figshare.29710445.v1

\bibitem{b31} N. Platt and P. Nayak, ``hvf-slm-v3-qwen: Fine-tuned Qwen2.5-7B for Maritime Intelligence,'' HuggingFace, 2025. [Online]. Available: https://huggingface.co/nolanplatt/hvf-slm-v3-qwen
\end{thebibliography}

\vspace{12pt}

\end{document}